\documentclass[letterpaper, 10 pt, conference]{ieeeconf}
\IEEEoverridecommandlockouts % for \thanks
\usepackage{graphicx} % for \includegraphics
\usepackage{booktabs}     
\usepackage[colorlinks, citecolor=red]{hyperref}
\usepackage{cite}
\usepackage{float}
\usepackage{comment}
\usepackage{arydshln}
\graphicspath{{./pdf/}} % load figure within subfolder ./pdf/
\DeclareGraphicsExtensions{.pdf}
\usepackage{algorithm}
\usepackage{algorithmic}

\renewcommand{\algorithmiccomment}[1]{\bgroup\hfill//~#1\egroup}
\usepackage[export]{adjustbox} % for \adjincludegraphics
\usepackage{threeparttable}
\usepackage{dsfont}
\usepackage[cmex10]{amsmath} % assumes amsmath package installed
\usepackage{colortbl}
\usepackage[short]{optidef} % for argmini environment

\usepackage{bm} % for \bm

\usepackage{epsfig} % for postscript graphics files

\usepackage{times} % assumes new font selection scheme installed

\usepackage{color} % for \color

\usepackage{amssymb}  % assumes amsmath package installed
\usepackage{multirow}

\newcommand{\mytilde}{\raise.17ex\hbox{$\scriptstyle\mathtt{‌​\sim}$}}

\usepackage{xspace}

\makeatletter
\DeclareRobustCommand\onedot{\futurelet\@let@token\@onedot}
\def\@onedot{\ifx\@let@token.\else.\null\fi\xspace}

\def\eg{\emph{e.g}\onedot} 
\def\ie{\emph{i.e}\onedot} 
 
 \def\vs{\emph{vs}\onedot}

\makeatother

\title{\LARGE \bf
	TraSeTR: Track-to-Segment Transformer with Contrastive Query for \\ Instance-level Instrument Segmentation in Robotic Surgery
}

\author{Zixu Zhao, Yueming Jin, and Pheng-Ann Heng% <-this % stops a space
	\thanks{This work was supported by the Key-Area Research and Development Program of Guangdong Province, China under Grant 2020B010165004, Hong Kong RGC TRS project T42-409/18-R, and a grant from the National Natural Science Foundation of China with Project No. U1813204.}
	
	\thanks{Z. Zhao and P. A. Heng are with the Department of Computer Science and Engineering, The Chinese University of Hong Kong, Hong Kong. P. A. Heng is also with Guangdong-Hong Kong-Macao Joint Laboratory of Human-Machine Intelligence-Synergy Systems,Shenzhen Institute of Advanced Technology, Chinese Academy of Sciences. Y. Jin is with the Department of Computer Science, University College London, UK.}
	\thanks{\emph{Corresponding author: Zixu Zhao (zxzhao@cse.cuhk.edu.hk) }}

}

\begin{document}

\maketitle
\thispagestyle{empty}
\pagestyle{empty}

\begin{abstract}

%Specifically, our transformer based method utilize each query embedding to be responsible for each tool, by predicting a binary mask associated with a single global class.
%Moreover, we introduce a new type of query (prior query) equipped with a new matching strategy (identity matching), to incorporate previous temporal knowledge to augment the query embedding by accurate tracking. 
%We further propose a contrastive training to reshape the query feature space, which largely alleviates the matching difficulty.

Surgical instrument segmentation -- in general a \textit{pixel classification} task -- is fundamentally crucial for promoting cognitive intelligence in robot-assisted surgery (RAS). 
However, previous methods are struggling with discriminating instrument types and instances.
To address above issues, we explore a \textit{mask classification} paradigm that produces per-segment predictions.  
We propose \textit{TraSeTR}, a novel Track-to-Segment Transformer that wisely exploits tracking cues to assist surgical instrument segmentation.
TraSeTR jointly reasons about the instrument type, location, and identity with instance-level predictions \ie, a set of class-bbox-mask pairs, by decoding query embeddings. 
Specifically, we introduce the prior query that encoded with previous temporal knowledge, to transfer tracking signals to current instances via identity matching. A contrastive query learning strategy is further applied to reshape the query feature space,  which greatly alleviates the tracking difficulty caused by large temporal variations.
%TraSeTR achieves coherent instance associations among frames in a tracking-by-attention mechanism that globally reasons about instrument type, location, and identity.
The effectiveness of our method is demonstrated with state-of-the-art instrument type segmentation results on three public datasets, including two RAS benchmarks from EndoVis Challenges and one cataract surgery dataset CaDIS.

\end{abstract}

%======================================================
\section{Introduction}
Robot-assisted surgery (RAS) has revolutionized the minimally invasive surgery by remarkably extending the dexterity and overall capability of surgeons. The robotic system controls the movement of surgical instruments, enabling efficient manipulation and vivid observation for many surgical tasks~\cite{shademan2016supervised, attanasio2020autonomous,qin2021learning}. 
Intelligent parsing of such instruments, \eg, identifying their types or positions, is highly desired for promoting cognitive assistance to surgeon perception~\cite{gao2021future}, operating workflow optimization~\cite{jin2021temporal}, and  skill assessment~\cite{reiley2009task,zia2018surgical}.
To this end, the instance-level semantic segmentation of instruments, which can separate instruments to different types,  is required as a fundamental task to support many downstream applications, such as tool pose estimation~\cite{sestini2021kinematic}, tracking~\cite{9380976}, trajectory prediction~\cite{osa2017online, toussaintco}, and even
\begin{figure}[!ht]
	\centering
	\includegraphics[width=0.48\textwidth]{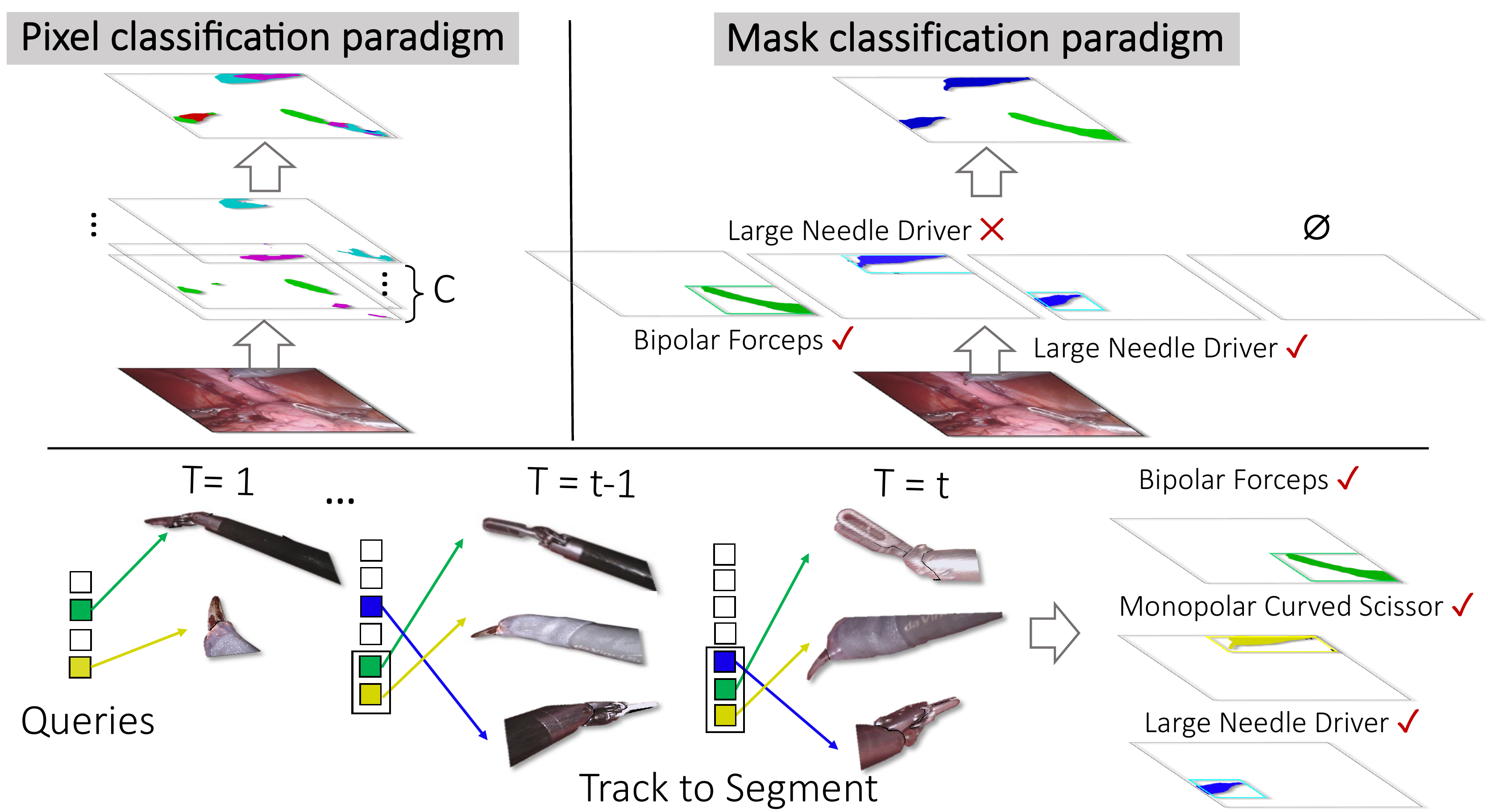}
	\vspace{-3mm}
	\caption{Two paradigms for surgical instrument type segmentation: (i) \textbf{pixel classification} predicts a single class for each pixel ($C$ classes in total).  (ii) \textbf{mask classification} predicts a set of binary masks and assigns a single class to each mask. To account for the large temporal variations of instruments, we exploit tracking cues with prior queries for video-level mask classification.}
	\vspace{-3mm}
	\label{fig:into}
\end{figure}
facilitate the surgical task automation~\cite{nagy2019dvrk,lu2021toward} for next generation of operating intelligence.

%Instance-level semantic segmentation of instruments, which can separate instruments to different types, is highly desired for promoting cognitive assistance to surgeon perception, operating workflow optimization and  skill assessment~\cite{reiley2009task,zia2018surgical}. Additionally, by densely identifying the positions of each tool, it can fundamentally support various RAS tasks, such as tool pose estimation~\cite{sestini2021kinematic}, tracking~\cite{9380976}, trajectory prediction~\cite{osa2017online}, and motion planning~\cite{toussaintco}, even can facilitate the surgical task automation~\cite{nagy2019dvrk} for next generation of operating intelligence.
%Intelligent parsing of such instruments, \eg, identifying their positions or types, is highly desired for promoting cognitive assistance and robot navigation~\cite{gao2021future}. To this end, the instance-level semantic segmentation of instruments is required as a fundamental task for developing RAS systems such as tool pose estimation~\cite{sestini2021kinematic}, tracking~\cite{9380976}, trajectory prediction~\cite{osa2017online}, and motion planning~\cite{toussaintco}, which in turn have applications ranging from operating theatre optimization to post-operative assessment~\cite{reiley2009task,zia2018surgical}. 

The similar instrument types, with small inter-class discrepancy, are challenging to recognize, especially in complex surgical scenes. Most of previous instrument segmentation methods follow a \textit{pixel classification} paradigm in which the deep learning model predicts the probability distribution over all classes for each pixel in a frame. The key idea of them~\cite{shvets2018automatic, jin2019incorporating, zhao2020learning, zhao2021one, zhao2021anchor} is to modify the neural network (\eg, U-Net~\cite{ronneberger2015u}) or  differentiate instrument types by exploring spatial or temporal cues, including depth maps~\cite{mohammed2019streoscennet}, pose estimation~\cite{kurmann2017simultaneous}, optical flows~\cite{jin2019incorporating}, and motion flows~\cite{zhao2020learning}. Nevertheless, as shown in Fig.~\ref{fig:into},  these solutions are struggling with the spatial class inconsistency problem, where one instrument may be assigned multiple instrument types.

An alternative paradigm -- \textit{mask classification} that predicts a set of binary masks, and each associated with a single class -- has been increasingly adopted for instance-level segmentation. In robotic surgery, ISINet~\cite{gonzalez2020isinet} takes the first step to predict a single class for each instrument segment based on Mask-RCNN~\cite{he2017mask}. One main challenge is to maintain the class consistency over time. The relabelling strategy in \cite{gonzalez2020isinet} takes into account the predictions of previous frames, but tends to misassign labels to similar instances due to large temporal variations  (see Fig.~\ref{fig:into}).
How to correctly perform  video-level mask classification for surgical instruments, to tolerate the intra-class variations across time, is crucial yet still remains unexplored.

\begin{figure*}[!ht]
	\centering
	\includegraphics[width=0.9\textwidth]{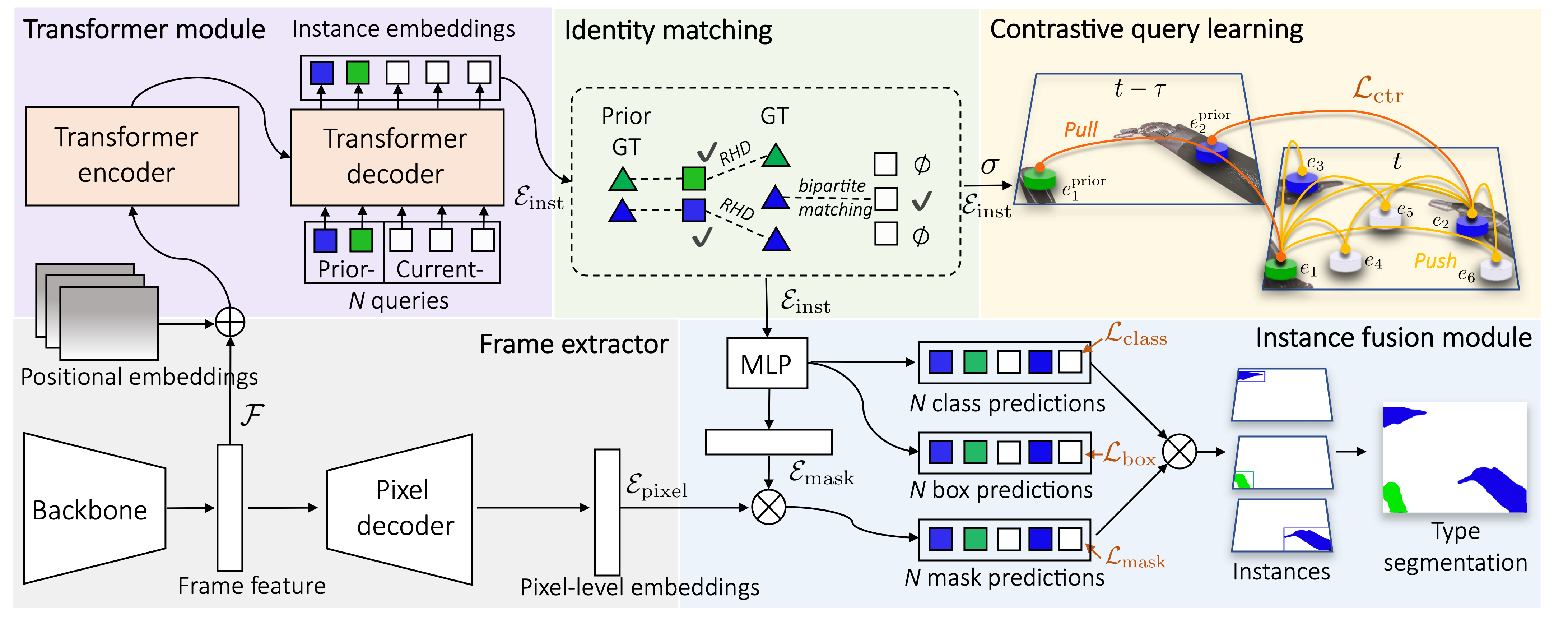}
	\vspace{-1mm}
	\caption{\textbf{TraSeTR overview.}  We use a backbone to extract frame feature $\mathcal{F}$ and a pixel decoder to produce pixel-level embeddings $\mathcal{E}_{\mathrm{pixel}}$. A transformer module computes from  $\mathcal{F}$ and $N$ queries to yield instance embeddings $\mathcal{E}_{\mathrm{inst}}$, which are combined with $\mathcal{E}_{\mathrm{pixel}}$ to output $N$ binary masks with bounding boxes and classes. TraSeTR performs instance tracking for surgical instrument segmentation. It is realized by a two-stage identity matching mechanism that allows the \textit{prior queries} to be precisely transformed to the current predictions, and the \textit{current queries} to infer newly-entered instances either initializing  prior queries for future frame. Based on the matching set $\sigma$, the contrastive query learning is applied to instance embeddings between time $t$ and $t-\tau$. 
	}
	\label{fig:method}
\end{figure*}

Recently, Transformers have shed light on mask classification by jointly reasoning about a number of query embeddings for instance predictions via the encoder-decoder attention mechanism~\cite{waswani2017attention}. The development of DETR~\cite{carion2020end} and its variants~\cite{zhu2020deformable} have widely demonstrated promising performance on object tracking~\cite{meinhardt2021trackformer} and instance segmentation~\cite{wang2021end,cheng2021per}.
Specifically in RAS scenarios, there emerge some works actively exploring transformer-based solutions for surgical phase recognition~\cite{gao2021trans}, tool detection~\cite{kondo2020lapformer}, and surgical scene reconstruction~\cite{long2021dssr}. 
All these successes motivate us to explore the potential of Transformer for discriminating instrument types in a way of
leveraging the temporal knowledge of dynamic instances.

In this paper, we propose \textit{TraSeTR}, a novel \textbf{tra}ck-to-\textbf{se}gment \textbf{tr}ansformer that dynamically integrates tracking cues to assist instance-level surgical instrument segmentation. Rooted in the mask classification paradigm, TraSeTR employs a Transformer module~\cite{waswani2017attention} to infer a set of instance predictions from  current frame features and query embeddings, each consisting of a class prediction, a bounding box prediction, and a binary mask prediction. The query embeddings are initialized as (i) \textit{prior queries} to encode prior knowledge of previous frame instances and (ii) 
\textit{current queries} to detect newly-appeared instances in a frame. Furthermore, they are learned to be temporally contrastive to tolerate the dynamic changes of instruments. 
TraSeTR performs  instance tracking via a two-stage identity matching between prediction set and ground-truth set such that the current instances can be precisely inferred from their corresponding prior queries. 
Our main contributions include:
\begin{itemize}

\item We propose a novel track-to-segment transformer to wisely discriminate both \textit{instances} and \textit{types} 
for accurate surgical instrument segmentation.

\item The keys to the success of TraSeTR are the new mechanisms of \textit{identity matching} and \textit{contrastive query learning}, which are carefully designed to track surgical instruments with large temporal variations.

\item We design a link-by-link inference strategy to infer instrument segments with dynamically changed prior queries in an auto-regressive fashion.

\item We extensively evaluated the proposed method on  two RAS benchmarks EndoVis17~\cite{allan20192017} and EndoVis18~\cite{allan20202018}, and a public eye surgery dataset CaDIS~\cite{grammatikopoulou2021cadis}. TraSeTR set new state-of-the-art results on instrument type segmentation tasks at a fast speed of 23 FPS.

%\item We also validated the instance-level tracking capability of our method. TraSeTR promisingly tracks each instrument entering or leaving the surgical scenes, which are crucial for 
\end{itemize}

%======================================================	
\section{Methods}
\label{METHODS}
TraSeTR is a  mask classification model for instance-level surgical instrument segmentation. In this section, we first describe how to formulate the track-to-segment problem. Then, we introduce the architecture and matching mechanism of the model. Finally, we illustrate the training and inference strategies which are specifically designed for TraSeTR.

\subsection{Track-to-Segment Formulation}
TraSeTR segments the instruments by (i) partitioning the frame into $N$ instances where $N$ is significantly larger than the real instrument number $\tilde{N}$, and (ii) merging the selected instances into one segmentation map over $C$ classes. The predictions of $N$ instances are a set of $N$ probability-bbox-mask pairs $z=\{(p_i, b_i, m_i)\}_{i=1}^N$, where the probability distribution $p_i$ contains an auxiliary ``no object'' label ($\varnothing$) to denote instances that do not correspond to any classes; $b_i\in[0,1]^4$ is the bounding box of the instance; and $m_i$ is the binary mask. To train the mask classification model, a matching set $\sigma$ between the  prediction set $z$ and the ground-truth set $\tilde{z}=\{(\tilde{c}_i, \tilde{b}_i, \tilde{m}_i)\}_{i=1}^{\tilde{N}}$ is required, where $\tilde{c}_i\in\{1,...,C\}$. 
The predictions that are not matched by $\sigma$ are assigned with the ground-truth label $\varnothing$.

For accurate video-level mask classification, TraSeTR incorporates tracking cues to assist segmentation by tracking instrument instances $\{z_1,...,z_k\}$ with identity $K= \{1,...k\}$. Specifically, TraSeTR initialize $N$ query embeddings for instance prediction. Among them, $N^{\mathrm{prior}}$ prior queries are initialized with the output embeddings of previous frame, and $N^{\mathrm{cur}}$ current queries are randomly initialized to learn to detect newly-entered instruments. Once the  matching set $\sigma$  assigns the predictions from prior queries to corresponding ground truths, the instances are successfully tracked.

\subsection{TraSeTR Architecture} 
The overall architecture of TraSeTR is simple and depicted in Fig.~\ref{fig:method}. 
We now describe three basic components that enable instance-level segmentation.

\textbf{Frame extractor.} Taking a frame  at time $t$ as input, the CNN backbone generates a feature map $\mathcal{F}\in \mathbb{R}^{\frac{H}{S} \times \frac{W}{S}  \times C_{\mathcal{F}}}$, where $C_{\mathcal{F}}$ is the  number of channels, $S$ is the sampling stride, and $H\times W$ is the frame size. The feature map is then upsampled via a decoder to produce pixel-level embeddings $\mathcal{E}_{\mathrm{pixel}}\in \mathbb{R}^{H W\times C_{\mathcal{E}}}$, where $ C_{\mathcal{E}}$ is the number of channels. 
%Besides, $\mathcal{F}$ is used for cross-attention in the transformer  to provide global visual information of the scene.

\textbf{Transformer module.} We employ a standard encoder-decoder Transformer~\cite{waswani2017attention} to compute from frame features $\mathcal{F}$ and $N$ query embeddings (including $N^{\mathrm{prior}}$ prior queries from time $t-\tau$ and $N^{\mathrm{cur}}$ current queries). We add $\mathcal{F}$ with positional embeddings as Transformer is permutation invariance. Thanks to the self- and cross-attention mechanisms in Transformer, its output, \ie, $N $ instance embeddings $\mathcal{E}_{\mathrm{inst}}\in \mathbb{R}^{N \times C_{\mathcal{Q}}}$ of dimension $C_{\mathcal{Q}}$, encode global information about all instruments appeared in the surgical scene. Note that the channel number of query embeddings is $C_{\mathcal{Q}}$ as well.

\textbf{Instance fusion module.} It first maps instance embeddings $\mathcal{E}_{\mathrm{inst}}$ to three types of predictions. For class prediction, we  apply a Multi-Layer Perceptron (MLP) followed by a softmax function to  yield class probability predictions $\{p_i\}_{i=1}^N$. For bounding box prediction, we use a MLP to produce $N$ bounding boxes $\{b_{i}\}_{i=1}^N$. For mask prediction, we convert $\mathcal{E}_{\mathrm{inst}}$ to  mask embeddings $\mathcal{E}_{\mathrm{mask}}\in \mathbb{R}^{N\times C_{\mathcal{Q}}}$ with a MLP, which are used to generate $N$ binary mask predictions $\{m_i\}_{i=1}^N$ via a dot production with pixel-level embeddings $\mathcal{E}_{\mathrm{pixel}}$, followed by a sigmoid activation. 
Finally, instances are fused to instrument type segmentation.
We assign each pixel $[h,w]$ to one of the probability-bbox-mask pair from the matching set $\sigma$ via $\mathrm{argmax}_{i\in \sigma}p_i(c_i)\cdot m_i[h, w]$. 
Here, $c_i$ is the predicted class $c_i = \mathrm{argmax}_{c\in\{1,...,C,\varnothing\}}p_i(c)$. 

\subsection{Tracking with Identity Matching}
\label{sec: matching}
TraSeTR infers a dynamic-size set of $N$ predictions in a single pass of the model, where $N$ is changing with the number of prior queries. To score the $N$ predictions with respect to the $\tilde{N}$ ground truths ($N>\tilde{N}$), we need to find a matching set $\sigma$ such that the $j^{\mathrm{th}}$ ground truth matches the prediction with index $\sigma(j)$. Essentially, the correct link between the prediction from prior query and the current ground truth can be regarded as a tracking task, which is crucial to enhance the temporal class consistency. 

However, tracking instruments is difficult because of the 
 large temporal variations (\eg, tool tips) caused by the zoom in and zoom out of endoscopic camera. The \textit{bipartite matching} used in DETR~\cite{carion2020end} and TrackFormer~\cite{meinhardt2021trackformer}, \ie, minimizing an assignment cost between the prediction $z_{\sigma(j)}$ and ground truth $\tilde{z}_j$,
$\mathcal{L}_{\mathrm{assign}}(z_{\sigma(j)}, \tilde{z}_j)= p_{\sigma(j)}(\tilde{c}_{j}) + \mathcal{L}_{\mathrm{box}}(b_{\sigma(j)}, \tilde{b}_{j})$, tends to cause invalid tracking by matching the prediction from current query to the ground truth. 

To correctly track instances, we propose a two-stage identity matching strategy. At the first stage, we search for a matching  subset $\sigma_1$ between predictions from prior queries and the current ground truths. A naive matching is possible if all instances belong to different classes. In this case, the $j^{\mathrm{th}}$ ground truth matches to $\sigma_1(j)^{\mathrm{th}}$ prediction if their classes are the same. A more general method can be achieved by employing the prior ground truths $\{\tilde{z}_i^{\mathrm{prior}}\}_{i=1}^{N^{\mathrm{prior}}}$ of these predictions for cost calculation, such that instrument identities can be explicitly associated with current ground truths $\{\tilde{z}_i\}_{i=1}^{\tilde{N}}$ via a matching cost defined as:
\begin{equation}
    \mathcal{L}_{\mathrm{match}}(z_{\sigma_1(j)}, \tilde{z}_{j}) = \mathds{1}_{\tilde{c}_{\sigma_1(j)}^{\mathrm{prior}} = \tilde{c}_{j}} |\tilde{x}^{\mathrm{prior}}_{\sigma_1(j)} - \tilde{x}_{j}|,
    \vspace{-1mm}
\end{equation}
where $\tilde{x}_j$ and $\tilde{x}^{\mathrm{prior}}_{\sigma_1(j)}$ are the horizontal coordinates of the $j^{\mathrm{th}}$ ground truth's center and $\sigma_1(j)^{\mathrm{th}}$ prior ground truth's center, derived from the bounding boxes $\tilde{b}_j$ and $\tilde{b}^{\mathrm{prior}}_{\sigma_1(j)}$, respectively. We then define the \textit{relative horizontal distance} (RHD) as  $|\tilde{x}^{\mathrm{prior}}_{\sigma_1(j)} - \tilde{x}_{j}|$ to distinguish instances, especially those of the same class, since the horizontal displacement of the same instance is smaller than that of the other. 
It can be inherently supported by the ``dual-arm distribution'', \ie, one instrument moves within the right-half scene, while the other (of the same class) within the left-half scene. Unlike the IoU-based bounding box loss~\cite{rezatofighi2019generalized}, RHD could still track the fast moving instruments in robotic surgery. 
The optimal subset then can be found by minimizing the matching cost:
\begin{equation}
    \hat{\sigma}_1 = \mathop{\arg\min}_{\sigma_1} \sum_{j} \mathcal{L}_{\mathrm{match}}(z_{\sigma_1(j)}, \tilde{z}_{j}),\; \small \sigma_1(j)\in [1, N^{\mathrm{prior}}].
\end{equation}
At the second stage, we search for a matching subset $\sigma_2$ between ground truths that are not matched by $\hat{\sigma}_1$ and predictions of current queries, to detect the newly-entered instance. Here, we optimize a bipartite matching-based assignment:
\begin{equation}
    \hat{\sigma}_2 = \mathop{\arg\min}_{\sigma_2} \sum_{j} \mathcal{L}_{\mathrm{assign}}(z_{\sigma_2(j)}, \tilde{z}_{j}),\; \small\sigma_2(j)\in (N^{\mathrm{prior}}, N].
\end{equation}
We assign a new identity (\ie, ID) for the newly-entered instance if its class appears in the video for the first time or there have already exist instances of the same class. Otherwise, it will be assigned a previous identity according to the class. Overall, the optimal matching set is a union of the subsets, \ie, $\sigma = \hat{\sigma}_1 + \hat{\sigma}_2$, where $\hat{\sigma}_2$ can be $\varnothing$.

\subsection{ Contrastive Learning-based Training}

Given the matching set $\sigma$, we encourage the model to transform the prior query to its current position, even with large temporal variations. Basically, we could compute the \textit{Hungarian loss} for all matched pairs, which are commonly used in prior works~\cite{carion2020end, meinhardt2021trackformer,cheng2021per}. This loss contains a 
negative log-likelihood for class prediction,  a bounding box loss and a mask loss for all instances:
\begin{equation}
\begin{aligned}
    \mathcal{L}_{\mathrm{Hung}}(z, \tilde{z})&= \sum_{j=1 }^{\tilde{N}}[-\mathrm{log}\; p_{\sigma(j)}(\tilde{c}_{j}) +  \mathcal{L}_{\mathrm{box}}(b_{\sigma(j)}, \tilde{b}_{j}) \\
     &+ \mathcal{L}_{\mathrm{mask}}(m_{\sigma(j)}, \tilde{m}_{j})].
\end{aligned}
\end{equation}
Here, we use the same $\mathcal{L}_{\mathrm{mask}}$ and $\mathcal{L}_{\mathrm{box}}$ as DETR~\cite{carion2020end}. Specifically, $\mathcal{L}_{\mathrm{mask}}$ is a linear combination of dice loss~\cite{sudre2017generalised} and focal loss~\cite{lin2017focal}. $\mathcal{L}_{\mathrm{box}}$ is a linear combination of $\ell_1$ loss and a generalized IoU loss $\mathcal{L}_{\mathrm{iou}}$~\cite{rezatofighi2019generalized}. 
%The \textit{Hungarian loss} induces joint constraints on the instance predictions.

\textbf{Contrastive query learning.} To further improve the tracking capability of TraSeTR based on the matching index $\sigma$, we push the model towards maximizing the query agreements on different temporal views of the instance. Let us denote $N$ instance embeddings $\mathcal{E}_{\mathrm{inst}}$ as $\{e_i\}_{i=1}^N$ and some of them are decoded from prior query embeddings.  Unlike previous video-level contrastive learning~\cite{sermanet2018time, zhao2021modelling}, our key idea (see Fig.~\ref{fig:method}) is to pull the instance embedding $e_i$ ($i\in \sigma$) with its prior query $e_i^{\mathrm{prior}}$, while pushing $e_i$ with the remaining instance embeddings $e_j$ where $c_i \neq c_j$. 
Formally, we construct two types of contrastive pairs, including the positive pairs $(e_i, e_i^{prior})$ and the negative pairs $(e_i, e_j)$, which allow us to learn contrastive query embeddings via a contrastive loss:
\begin{equation}
    \mathcal{L}_{\mathrm{ctr}} = - \frac{1}{|\sigma|}\mathrm{log} \sum_{i\in \sigma} \frac{ \phi(e_i, e_i^{prior})}{ \phi(e_i, e_i^{prior}) + \sum_{j} \phi(e_i, e_j)},
\label{eq:1}
\end{equation}
where $\phi(\cdot, \cdot)$ is a similarity function and can be achieved by a dot production between two embeddings. In Eq.~\eqref{eq:1}, we omit the calculation of instance embeddings that are matched by $\sigma$ but not have prior embeddings, \ie, those newly-entered instances. We now define the overall training loss as $ \mathcal{L}_{\mathrm{Hung}} + \lambda_{\mathrm{ctr}}\mathcal{L}_{\mathrm{ctr}}$, where  $\lambda_{\mathrm{ctr}}$ is a balancing weight.

\subsection{Link-by-link Inference}
We describe a link-by-link inference procedure specifically designed for instance-level instrument segmentation. At the beginning, TraSeTR predicts all the instruments that appear in the first frame with the identity $K_1 = \{1,...,k\}$ being a subset of all $K$ in the video sequence. It only decodes $N^{\mathrm{cur}}$ instance embeddings and then select $k$ instances whose classification scores are above $\tau_d$. The $k$ output embeddings are used to initialize the prior queries for the next frame.
At time $t$, TraSeTR outputs $N^{\mathrm{cur}} + N^{\mathrm{prior}}$ instance embeddings. Apart from detecting newly-appeared instruments with classification threshold $\tau_d$, TraSeTR tracks prior instances whose classification scores are above $\tau_t$. 
Note that $N^{\mathrm{prior}}$ changes between frames as prior queries are removed or new instances are detected. We remove prior queries if their classification scores drop below $\tau_t$ for more than 50 time steps. The time tolerance allows TraSeTR to infer 
an instance with multiple prior queries collected from different time steps, which provides  the long-range temporal information of one certain instrument.

%======================================================

\section{Experiments and Results}
\label{EXPERIMENTS}

We evaluated TraSeTR's performance on instrument type segmentation using three public datasets~\cite{allan20192017,allan20202018,grammatikopoulou2021cadis}.

\subsection{Datasets and Evaluation Metrics}

 \textbf{EndoVis17.} The EndoVis17 dataset~\cite{allan20192017}, a benchmark of instrument type segmentation, contains 8 robot-assisted surgery videos recorded from da Vinci Xi Surgical System. We used the instance annotations generated by~\cite{gonzalez2020isinet}.
%including Bipolar Forceps, Prograsp Forceps, Large Needle Driver, Vessel Sealer, Grasping Retractor, Monopolar Curved Scissors, and Ultrasound Probe. 

\textbf{EndoVis18.} The EndoVis18 dataset~\cite{allan20202018} provides 
15 videos of different porcine procedures acquired by da Vinci Xi Surgical System, and corresponding semantic annotations of the whole scene. The instruments are annotated with their parts (\textit{shaft, wrist} and \textit{jaws}). To distinguish among instrument types, we followed prior work~\cite{gonzalez2020isinet} to generate the additional instance annotations for 7 instrument types. 

\textbf{CaDIS.} The CaDIS dataset~\cite{grammatikopoulou2021cadis} includes 25 surgical videos recording cataract surgery by an OPMI Lumera T microscope. We used the semantic annotations of instruments and converted them to instance annotations by extracting each instrument from the scene and assigning it one of the 10 instrument types in Table~\ref{tab:result3}. 
Our annotations can be transformed to the MS-COCO standard dataset format. 

As the ground-truth instance ID is not provided, we only assess the segmentation quality following prior works~\cite{shvets2018automatic, jin2019incorporating, zhao2020learning,gonzalez2020isinet,grammatikopoulou2021cadis}. Specifically, we adopted two commonly used metrics including mean intersection-over-union (mIoU) and Dice coefficient (Dice) that only consider the classes presented in a frame. For fair comparisons, EndoVis17 was evaluated by 4-fold cross-validation using the standard folds described in~\cite{shvets2018automatic}. EndoVis18 and CaDIS were evaluated using the same data splitting as~\cite{gonzalez2020isinet} and~\cite{grammatikopoulou2021cadis}, respectively. 

\begin{table*}[!ht]
\caption{Instrument Type Segmentation Results of Different Methods on EndoVis17 and EndoVis18 datasets (7 classes). }
\label{tab:result}
\centering
\resizebox{\textwidth}{!}{
\begin{threeparttable}
\begin{tabular}{cccccccccccc}
\hline
 \multirow{2}{*}{Dataset} &
  \multirow{2}{*}{Method} &
  \multirow{2}{*}{Bbox} &
  \multirow{2}{*}{mIoU} &
  \multirow{2}{*}{Dice} & 
  \multicolumn{7}{c}{Instrument classes (mIoU)} \\ \cline{6-12} 
                       &    &            &       &       & BF    & PF    & LND   & VS / SI & GR / CA & MCS   & UP    \\ \hline
\multicolumn{1}{l}{\multirow{7}{*}{EndoVis17}} &
  TernausNet~\cite{shvets2018automatic} & & 35.3 & 44.9 & 13.3 &  12.4 & 20.5 & 6.0 & 1.1 &  1.0 & 16.8 \\
 \multicolumn{1}{l}{} &MF-TAPNet~\cite{jin2019incorporating} &  &  37.4 & 48.0 & 16.4 &  14.1 &  19.0 &  8.1 & 0.3 &4.1 & 13.4 \\
\multicolumn{1}{l}{} & Dual-MF~\cite{zhao2020learning}& &45.8 &  56.1 &34.4 &21.5 &64.3 &24.1& 0.8 &17.9 &\textbf{21.8} \\\cdashline{2-12}
\multicolumn{1}{l}{} & DETR~\cite{carion2020end}&$\checkmark$  &53.1 &58.0&  36.5 &  37.2 &  54.5 &  24.2 &  0.6 &  23.3 & 11.3 \\
\multicolumn{1}{l}{} & TrackFormer~\cite{meinhardt2021trackformer}&$\checkmark$  &54.9 &59.7&  37.6 &  38.0 &  53.1 &  25.5 &  2.8 &  24.6 & 15.7 \\
\multicolumn{1}{l}{} & ISINet~\cite{gonzalez2020isinet}&$\checkmark$  &55.6 &62.8&  38.7 &  38.5 &  50.1 &  27.4 &  2.0 &  28.7 & 12.6 \\\cdashline{2-12}
%\multicolumn{1}{l}{} & MaskFormer~\cite{strudel2021segmenter}&  &50.2 &59.5&  40.2 &  33.6 &  60.1 &  21.0 &  5.4 &  17.0 & 13.8 \\\cdashline{2-12}
\multicolumn{1}{l}{} &  \textbf{TraSeTR} (ours) &$\checkmark$ &  \textbf{60.4} \textcolor{red}{(+4.8)} & \textbf{65.2} \textcolor{red}{(+2.4)} &  \textbf{45.2} & \textbf{56.7} &  \textbf{55.8} & \textbf{38.9} &  \textbf{11.4} &  \textbf{31.3} &  18.2 \\ \hline
\multirow{7}{*}{EndoVis18} & TernausNet~\cite{shvets2018automatic} & &  46.2 & 53.2 &  44.2 &  4.7 &  0.0 &  0.0 &  0.0 & 50.4 & 0.0 \\
  &MF-TAPNet~\cite{jin2019incorporating} &  &  67.9 & 72.5 & 69.2 &  6.1 & 11.7 &  14.0 &
  0.9 &  70.2 &  0.6 \\
 &
  Dual-MF~\cite{zhao2020learning} & &
  70.4 &
  76.9 &
  74.1 &
  6.8 &
  46.0 &
  30.1 &
  7.6 &
  80.9 &
  0.1 \\\cdashline{2-12}
  & DETR~\cite{carion2020end}&$\checkmark$  &68.0 &72.5&  70.3 &  15.9 & 31.6 &  16.7 &  0.9 &  80.2 & 0.0 \\
  & TrackFormer~\cite{meinhardt2021trackformer}&$\checkmark$  &71.1 &77.3 &  75.8 & 20.1 &  38.5 & 30.6 &  4.8 & 82.5 & 1.5 \\
  %& MaskFormer~\cite{strudel2021segmenter}&  &71.7 &77.9&  72.5 & 26.5 &  41.2 &  35.3 &  3.4 &  81.9 & 2.5 \\\cdashline{2-12}
  &
  ISINet~\cite{gonzalez2020isinet} &$\checkmark$ &  
  73.0 &
  78.3 &
  73.8 &
  48.9 &
  31.0 &
  37.7 &
  0.0 &
  \textbf{88.2} &
  2.2 \\\cdashline{2-12}
  &\textbf{TraSeTR} (ours) &$\checkmark$ &\textbf{76.2} \textcolor{red}{(+3.2)}
   &\textbf{81.0} \textcolor{red}{(+2.7)}
   &\textbf{76.3}
   &\textbf{53.3}
   &\textbf{46.5}
   &\textbf{40.6}
   &\textbf{13.9}
   &86.3
   &\textbf{17.5}
   \\ \hline
\end{tabular}
\begin{tablenotes}
\footnotesize
\item Instrument classes include: Bipolar Forceps (BF), Prograsp Forceps (PF), Large Needle Driver (LND), Vessel Sealer (VS), Suction Instrument (SI), Grasping Retractor (GR), Clip Applier (CA), Monopolar Curved Scissors (MCS), and Ultrasound Probe (UP). 
\end{tablenotes}
\end{threeparttable}
}
\end{table*}

\subsection{Implementation Details}

\textbf{Model settings.} TraSeTR is compatible with any backbone architecture. In this work, we use the ResNet-50~\cite{he2016deep} backbone. The Transformer consists of 6 encoder and 6 decoder layers with 8 attention heads. As TraSeTR predicts 1$\sim$4 instruments for each frame, the query number $N^{\mathrm{cur}}$ is 20 (see query number  ablation in supplementary video).

\textbf{Training.} The time interval $\tau$ between prior frame and current frame is in the range of [1, 10]. Our model is implemented in Pytorch and trained with a NVIDIA Titan Xp GPU. We initialize the backbone with pretrained weights on COCO~\cite{lin2014microsoft}. The initial learning rates of Transformer and backbone are $1e\!-\!4$ and $1e\!-\!5$, which will be multiplied by 0.1 after 50 epochs. We use the same balancing weights as DETR~\cite{carion2020end} for $\mathcal{L}_{\mathrm{Hung}}$. The hyper-parameter $\lambda_{\mathrm{ctr}}$ is set as 0.2.
To increase model robustness, we augment query embeddings by adding false negatives with a probability of 0.4 and false positives with a probability of 0.1, following~\cite{meinhardt2021trackformer}.

\textbf{Inference.} During inference, we set the detection threshold $\tau_{d}$ as 0.9. To tolerate  large temporal variations of instruments, we set the track threshold $\tau_{t}$ as 0.6. We also apply non-maximum suppression (NMS) with a high IoU threshold of 0.9 to filter out overlapped instances. The inference speed could be 23 FPS without extra acceleration.

\subsection{Main Results}
\begin{figure*}[!ht]
	\centering
	\includegraphics[width=0.92\textwidth]{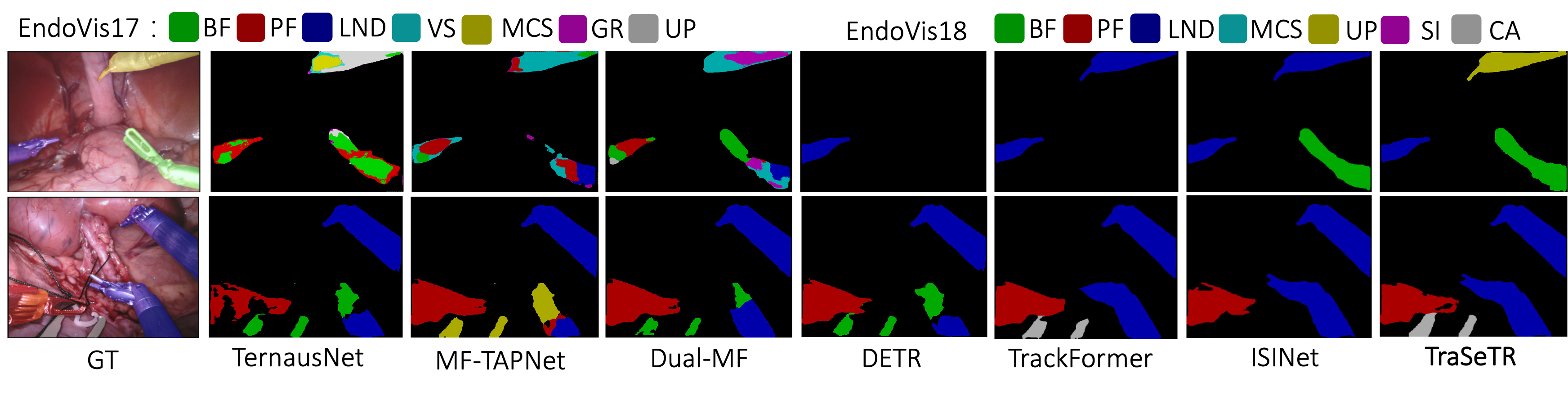}
	\vspace{-5mm}
	\caption{\textbf{Qualitative comparisons} of TernausNet~\cite{shvets2018automatic}, MF-TAPNet~\cite{jin2019incorporating}, Dual-MF~\cite{zhao2020learning}, DETR~\cite{zhu2020deformable}, TrackFormer~\cite{meinhardt2021trackformer}, ISINet~\cite{gonzalez2020isinet}, and our TraSeTR on EndoVis17 (top) and EndoVis18 (bottom) videos. Each color represents one instrument type. More results can be found in supplementary video.}
	\vspace{-2mm}
	\label{fig:result}
\end{figure*}

\subsubsection{EndoVis 17\&18} In Table~\ref{tab:result}, we compare our TraSeTR with (i) \textit{pixel classification approaches}~\cite{shvets2018automatic, jin2019incorporating, zhao2020learning} and (ii) \textit{mask classification approaches}~\cite{zhu2020deformable, meinhardt2021trackformer, gonzalez2020isinet} for instrument type segmentation. For DETR~\cite{zhu2020deformable} and TrackFormer~\cite{meinhardt2021trackformer}, we use the source code provided by the authors to train the models. Other results are reported from the original papers. For EndoVis17, TraSeTR outperforms Dual-MF~\cite{zhao2020learning} by 14.6\% mIoU and 9.1\% Dice, indicating that mask classification formulation has great potential for instrument type segmentation.  Compared with transformer-based approaches~\cite{zhu2020deformable, meinhardt2021trackformer}, TraSeTR shows great improvements by tracking instances such that prior queries can be leveraged to infer current instances. TraSeTR is also superior to ISINet ~\cite{gonzalez2020isinet}, achieving a new state-of-the-art of 60.4\% mIoU and 65.2\% Dice. For EndoVis18, TraSeTR still outperforms the prior state-of-the-art~\cite{gonzalez2020isinet} by 3.2\% mIoU and 2.7\% Dice. In particular, the improvements of some types, \eg, Clip Applier and Ultrasound Probe, are more than 10\% mIoU. Fig.~\ref{fig:result} shows the qualitative comparisons. As expected, TraSeTR  maintains both  spatial and temporal class consistency of instruments, while other methods fail to do so.

\subsubsection{CaDIS} Table~\ref{tab:result3} compares the type segmentation results of TraSeTR and three strong baselines reported in~\cite{grammatikopoulou2021cadis}, including DeepLabV3+~\cite{chen2018encoder}, 
UPerNet~\cite{xiao2018unified}, and HRNetV2~\cite{wang2020deep}.
Observe that TraSeTR achieves the best overall results of 69.9\% mIoU. For some certain types, \eg, Ph. Handpiece and I/A Handpiece, the similar tool tips could be better distinguished by the high-resolution representations of HRNetV2~\cite{wang2020deep}. But promisingly,
TraSeTR peaks the segmentation performance of 7 instrument types. This result suggests that our method is robust to various surgical instruments and surgical scenes.
%DeepLabV3+ leverages long-range contextual information using atrous convolutions with different dilation rates. UperNet uses a Feature Pyramid Network to extract multi-scale features and a pyramid pooling module to incorporate both global and local contextual information. HRNetV2 produces strong high-resolution feature representations by aggregating features from all the parallel convolutions.

\begin{table}[!ht]
\caption{Instrument Type Segmentation Results (mIoU) of Different Methods on CaDIS Dataset (10 classes). }
\label{tab:result3}
\centering
\resizebox{0.49\textwidth}{!}{
\begin{tabular}{lcccc}
\hline
\begin{tabular}[c]{@{}c@{}}Instrument\\ classes\end{tabular} &
  \begin{tabular}[c]{@{}c@{}}DeepLabV3+\\ \cite{chen2018encoder}\end{tabular} &
  \begin{tabular}[c]{@{}c@{}}UPerNet\\ \cite{xiao2018unified}\end{tabular} &
  \begin{tabular}[c]{@{}c@{}}HRNetV2\\ \cite{wang2020deep}\end{tabular} &
  \begin{tabular}[c]{@{}c@{}}\textbf{TraSeTR}\\(ours) \end{tabular}\\ \hline
Cannula          & 48.9 & 50.0 & 49.5 & \textbf{53.1} \\
Cap. Cystotome   & 55.7 & 54.5 & 61.7 &  \textbf{64.4}\\
Tissue Forceps   & 70.0 & 74.0 & 78.0 &  \textbf{79.2}\\
Primary Knife   & 86.1 & 89.5 & 89.3 &  \textbf{92.3}\\
Ph. Handpiece    & 75.0 & 77.6 & \textbf{77.9} &  76.2\\
Lens Injector    & 78.5 & 81.0 & 82.8 &  \textbf{87.1}\\
I/A Handpiece   & 74.0 & 73.6 & \textbf{75.3} &  71.3\\
Secondary Knife  & 69.0 & 68.2 & 79.5 &  \textbf{82.7}\\
Micromanipulator & 59.3 & 63.6 & 64.4 &  \textbf{66.3}\\
Cap. Forceps     & \textbf{28.9} & 23.0 & 27.2 &  26.1\\\hline
Total              & 64.6 & 65.5 & 68.6 &  \textbf{69.9}\\\hline
\end{tabular}
}
\end{table}

\subsubsection{Instance-level Tracking} 
\begin{figure*}[!ht]
	\centering
	\includegraphics[width=0.91\textwidth]{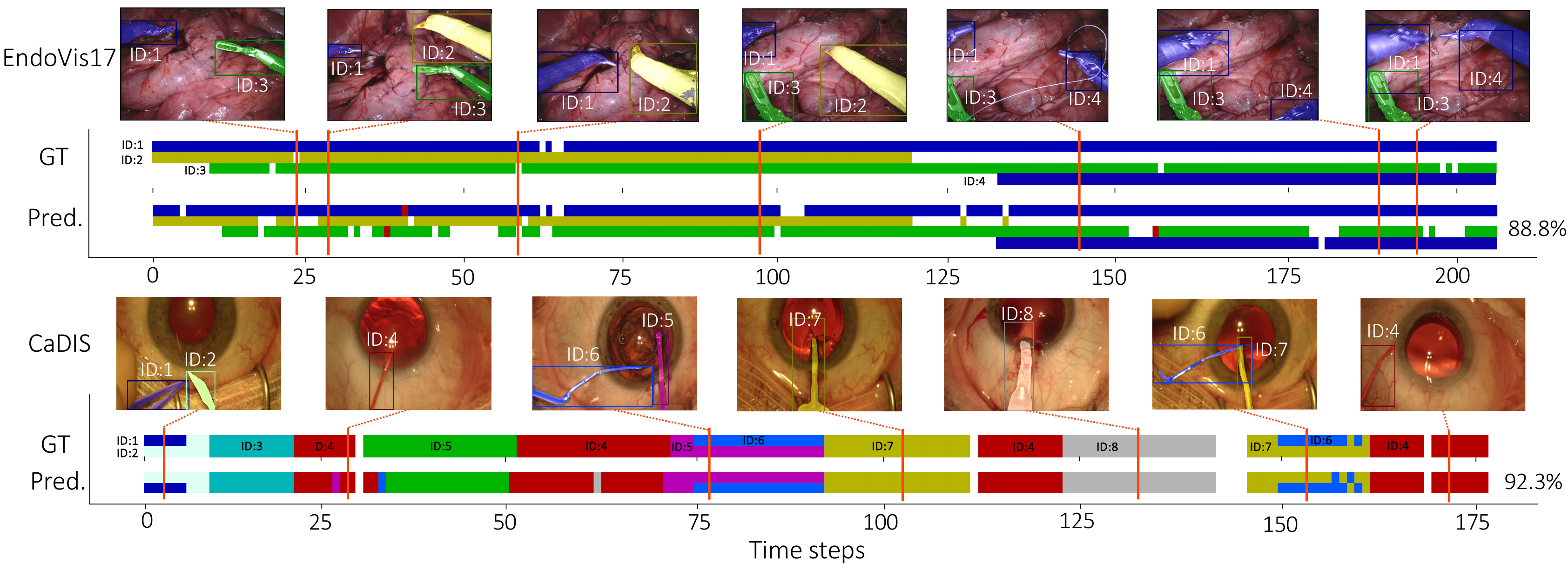}
	\vspace{-3mm}
    	\caption{ \textbf{Visualization of instance-level tracking} on EndoVis17 and CaDIS videos. The color bar represents the time steps that one instrument appearing in the scene. Different color indicates different instrument types. Besides the tracking process, we also show the instrument type segmentation results of intermediate frames.
    	TraSeTR achieves 88.8\% and 92.3\% tracking accuracy (\ie, how many instances being correctly tracked) on two example videos.} 
	\label{fig:track}
\end{figure*}

Fig.~\ref{fig:track} visualizes the tracking process of our method. Each color bar represents an instrument with a specific identity (ID) and class. The length of the bar indicates the time span of the instrument showing in the video. Compared with ground truth, TraSeTR correctly tracks  instruments that newly-appear or re-appear in the scenes. More importantly, it assigns the instance ID to each of them, thereby achieving satisfying instance-level tracking.

\subsection{Ablation Studies}
\subsubsection{Bipartite matching \vs identity matching} In Table~\ref{tab:b} (a), we verify that the main gains of TraSeTR come from tracking instances with identity matching. We start by comparing (i) TraSeTR uses bipartite matching, and (ii) TraSeTR uses identity matching. We report the \textit{mean tracking rate} of two methods at training time, \ie,  how many predictions from prior queries can be matched to the current ground truths. Only 27.3\% instances can be successfully tracked via bipartite matching, which means that the model cannot always use prior queries for prediction. On the contrary, our identity matching tracks all instances by associating their identities at training time. As a result, it leads to 3.5\% mIoU and 2.1\% Dice improvements on EndoVis17 dataset, suggesting the necessity of fully exploiting temporal information to discriminate instrument types.

\begin{table}[!ht]
\caption{Ablation studies of TraSeTR on EndoVis17 Dataset.}
\label{tab:b}
\centering
\resizebox{0.48\textwidth}{!}{
\begin{tabular}{c|ccc|cc}
\hline
\multicolumn{1}{c|}{(a)}                  & \multicolumn{3}{c|}{Mean tracking rate (at training)}   & mIoU                  & Dice                  \\ \hline
Bipartite matching    & \multicolumn{3}{c|}{27.3\% } &  56.9    &   63.1   \\
Identity matching     & \multicolumn{3}{c|}{100\% } &\textbf{60.4}    &   \textbf{65.2}  \\ \hline\hline
\multicolumn{1}{c|}{\multirow{2}{*}{(b)}} & \multicolumn{3}{c|}{Query Embeddings} & \multirow{2}{*}{mIoU} & \multirow{2}{*}{Dice} \\
\multicolumn{1}{c|}{} & Current     & Prior     & Contrastive     &      &      \\ \hline
TraSeTR-NT            &   $\checkmark$    &       &       & 54.6 & 62.0 \\
TraSeTR-NC            &    $\checkmark$   &    $\checkmark$   &       & 59.6 & 64.7 \\
TraSeTR               &   $\checkmark$    &   $\checkmark$    &  $\checkmark$     & \textbf{60.4} & \textbf{65.2} \\ \hline
\end{tabular}
}
\end{table}

\subsubsection{Types of Query embeddings} 
\begin{figure}[!ht]
	\centering
	\includegraphics[width=0.41\textwidth]{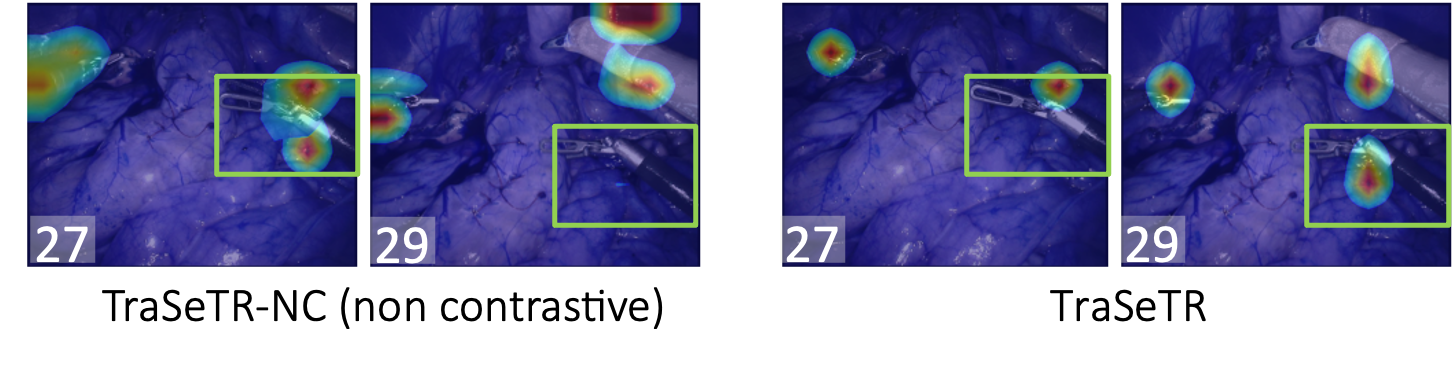}
	\vspace{-3mm}
	\caption{Attention maps of TraSeTR-NC and TraSeTR. We visualize instance embeddings whose indexes are in the matching set $\sigma$ using the projection algorithm~\cite{meinhardt2021trackformer}.
	The green box indicates the largely changed instrument.} 
	\label{abl:ctr}
\end{figure}

Table~\ref{tab:b} (b) analyzes the different types of query embeddings in TraSeTR. We implement three configurations: (i) \textit{TraSeTR-No Tracking} (NT): TraSeTR with current queries only, and trained with Hungarian loss $\mathcal{L}_{\mathrm{Hung}}$; (ii) \textit{TraSeTR-Non Contrastive} (NC): TraSeTR with current and prior queries, and trained with $\mathcal{L}_{\mathrm{Hung}}$; (iii) \textit{TraSeTR}: TraSeTR with contrastive queries, and trained with a combination of $\mathcal{L}_{\mathrm{Hung}}$ and contrastive loss  $\mathcal{L}_{\mathrm{ctr}}$. TraSeTR-NT achieves instance-level instrument segmentation but tends to predict wrong classes for some segments.
Adding  prior queries  alleviates this issue as the temporal information can be explicitly leveraged. The contrastive query embeddings further improve the model's discrimination capability of largely changed instruments, peaking the segmentation results on two metrics. As shown in Fig.~\ref{abl:ctr}, contrastive query learning inherently strengthens the encoder-decoder attention mechanism in TraSeTR, such that the instance attention regions can be precisely found.

\begin{comment}
\subsubsection{Number of instance queries}
Table~\ref{tab:query} shows the results of TraSeTR trained with a varying number of instance queries on three datasets with different number of instrument types. The model with 20 instance queries consistently achieves the best segmentation performance across all datasets, which suggests that we do not need to adjust the number of instance queries according to different dataset. As the number of instruments appeared in the scene is relatively small (1$\sim$4 in most cases), it is not necessary to use more than 50 instance queries to detect the instruments. For example, the model with 100 instance queries suffers from a  performance decay due to the minor inter-query variations. 

\begin{table}[!ht]
\caption{Instrument Type Segmentation Results Using Different Number of Instance Queries. }
\label{tab:query}
\centering
\resizebox{0.45\textwidth}{!}{
\begin{tabular}{c|cc|cc|cc}
\hline
\multirow{2}{*}{\begin{tabular}[c]{@{}c@{}}Number\\ of queries\end{tabular}} & \multicolumn{2}{c|}{EndoVis17} & \multicolumn{2}{c|}{EndoVis18} & \multicolumn{2}{c}{CaDIS} \\
    & mIoU & Dice & mIoU & Dice & mIoU & Dice \\ \hline
10  & 58.3 & 62.5 & 77.0 & 80.1 & 68.8 & 70.6 \\
20  & \textbf{60.4} & \textbf{65.2} & \textbf{78.2} & \textbf{82.5} & \textbf{69.5} & \textbf{71.3} \\
50  & \textbf{60.4} & 65.0 & 77.6 & 81.8 & 69.3 & \textbf{71.3} \\
100 & 53.2 & 55.4 & 72.6 & 74.0 & 65.3 & 66.1 \\ \hline
\end{tabular}
}
\end{table}
\end{comment}

%======================================================	
\section{Conclusion And Future Work}
\label{CONCLUSIONS}	

This paper presents a novel transformer-based mask classification approach to dynamically track instances in robotic surgical video for accurate semantic segmentation of instruments. 
Our method addresses the difficulties in this task, \ie, most notably the small inter-class discrepancy and large intra-class variations of instruments, by fully leveraging the set prediction mechanism in the designed transformer to produce per-instance predictions, and a identity matching strategy to incorporate tracking cues.
TraSeTR was evaluated on three public datasets, including two  RAS datasets and one cataract surgery dataset that contains different instrument types and surgical techniques performed in diverse platforms. 
TraSeTR outperforms the  state-of-the-art performance by up to 5\% mIoU and promisingly tracks the positions of instruments entering or leaving the scene. 
The improvements can facilitate the real-world RAS task automation, such as suturing and dissection, which greatly benefit from instance-level perception.
%The improvements can facilitate the real-world RAS task automation, which greatly benefit from instance-level perception, for example, taking the relatively simple tasks of suturing and dissection as the first step. 
%The improvements are significant for real-world RAS tasks such as suturing and dissection, which greatly benefit from instance-level perception. 
Ablation studies demonstrated the effectiveness of our transformer design and the necessity of contrastive query inductions to tolerate temporal variations. 
We plan to further investigate the alternative guidance for instance-level instrument segmentation. One potential direction is to integrate the multi-modal data, \eg, the kinematics data (the position, velocity of the tool tips) from the robotic systems, into the flexible transformer architecture. We will also extend TraSeTR into a 
unified framework to generate instrument trajectory maps online, which can be applied to downstream scenarios such as robot motion planning in RAS.

%\addtolen\mathrm{gt}h{\textheight}{-12cm}   % This command serves to balance the column len\mathrm{gt}hs
		
%\section*{ACKNOWLEDGMENT}

%The authors would like to thank Colin Lea for sharing raw features of JIGSAWS dataset. 
	
\clearpage
	
\bibliographystyle{IEEEtran}
\bibliography{mybibfile}

\end{document}